\documentclass[letterpaper, 10 pt, conference]{ieeeconf}  

\IEEEoverridecommandlockouts                              

\usepackage{graphics} 
\usepackage{epsfig} 
\usepackage{mathptmx} 
\usepackage{mathtools}
\usepackage{times} 
\usepackage{amsmath} 
\usepackage{amssymb}  

\usepackage{xcolor}
\usepackage{cite}
\usepackage[hidelinks]{hyperref}    
\title{\LARGE \bf
Grasp2Twist: Learning Bimanual Dexterous Jar Opening by Reinforcement Learning}

\author{
Mo Xu$^{1}$,
Yunfu Deng$^{1}$,
Jianuo Wang$^{2}$,
Josiah Hanna$^{1,\dagger}$,
Bilge Mutlu$^{1,\dagger}$%
\thanks{$^{1}$University of Wisconsin--Madison, Madison, WI, USA.}
\thanks{$^{2}$Independent Researcher.}
\thanks{$^{\dagger}$Josiah Hanna and Bilge Mutlu are equal senior authors.}
\thanks{This work has been submitted to the IEEE for possible publication.
Copyright may be transferred without notice, after which this version may no longer be accessible.}%
}

\begin{document}

\maketitle
\thispagestyle{empty}
\pagestyle{empty}

\begin{abstract}

This paper presents Grasp2Twist, a bimanual dexterous manipulation system that learns to grasp and twist open jar lids using reinforcement learning. 
Learning this task raises three challenges: learning a unified policy for a multi-stage task, sustaining lid twisting, and sim-to-real transfer.
To address the first challenge, we introduce a continuous enclosure measure to guide grasp formation and a binary enclosure indicator to guide the grasp-to-twist transition for unified policy learning. We derive both from the geometric relationship between the object center and the convex hull formed by the hand's palm and fingertips.
Kinematic constraints limit how far the hand can rotate the lid with fixed contacts, so sustained twisting requires finger contact reconfiguration. We use a three-stage curriculum to facilitate exploration of these contact changes and also improve robustness for sim-to-real transfer.
With our approach, the learned policy demonstrates finger gaiting, reconfiguring hand-object contacts to sustain lid rotation.
It transfers zero-shot to the physical system and achieves an 88\% task success rate across six household containers, including peanut-butter, vitamin, and instant-coffee jars.
Ablations further validate the roles of the geometric enclosure in grasp formation and the curriculum in contact-reconfiguration exploration.

\end{abstract}

\section{Introduction}
Bimanual dexterous manipulation enables robots to use two multi-fingered hands in complementary roles, stabilizing an object with one hand while manipulating it with the other. 
However, coordinating both arms and hands is challenging due to the high dimensionality of the control problem and the disturbances transmitted between the hands through the shared object.
Jar opening is one task that exemplifies the need for coordination: one hand grasps and holds the jar body while the other grasps and twists the lid until the threads disengage. 
Learning this task with a single policy requires coordinating grasp acquisition with subsequent lid twisting and jar stabilization. In this paper, we study the challenges of bimanual dexterous manipulation using the task of jar opening.

Bimanual dexterous jar opening is a contact-rich task. Twisting with fixed finger contacts eventually exhausts the hand's available motion range~\cite{morgan2022complex}, limiting further lid rotation. Sustained lid twisting therefore requires reconfiguring these contacts.
Transferring these contact-reconfiguration behaviors from simulation to hardware is also challenging because discrepancies in actuation and contact dynamics can disrupt the contact changes and destabilize the grasps.

\begin{figure}[!t]
    \centering
    \includegraphics[width=\linewidth]{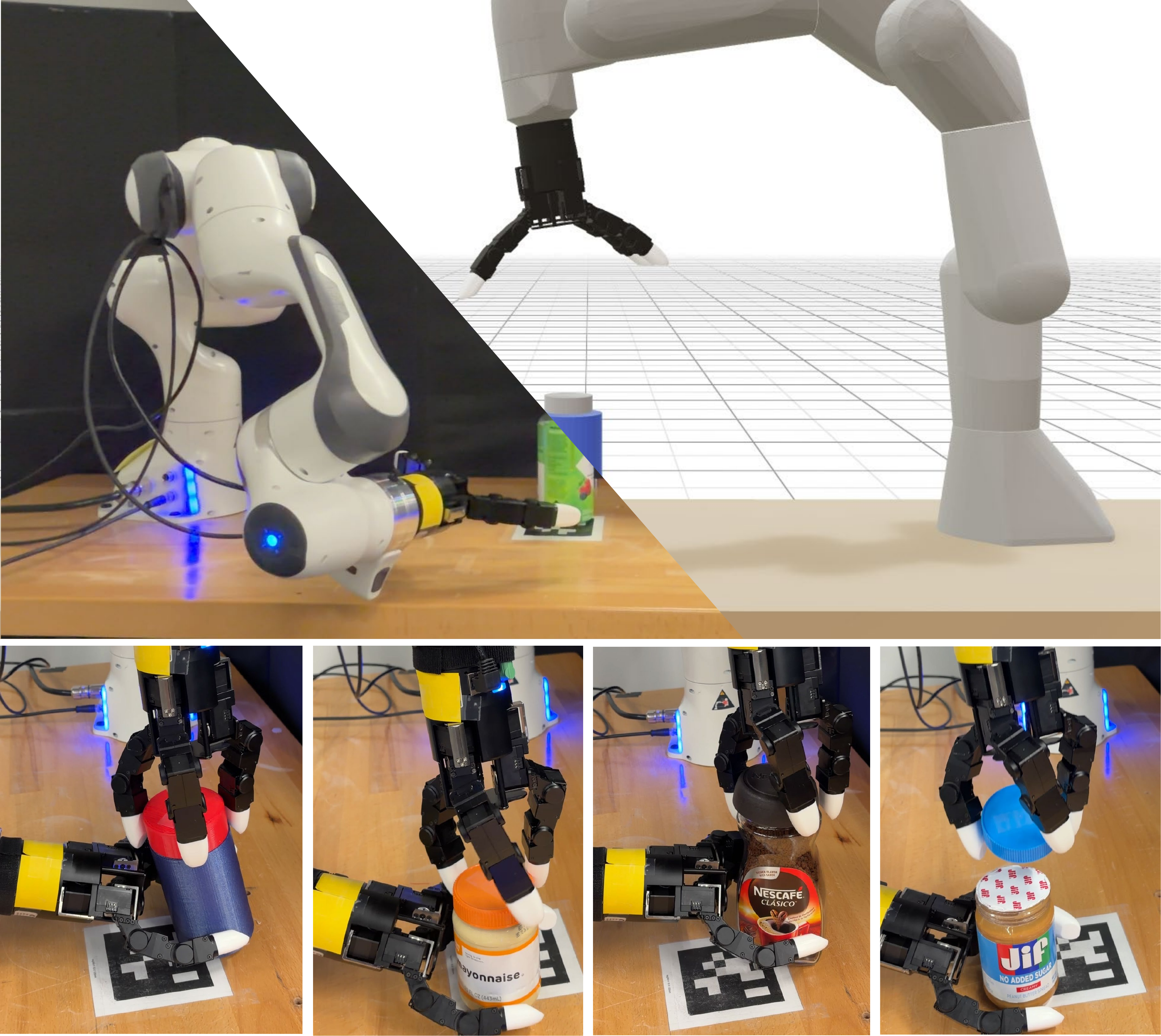}
    \caption{Real-world and simulated bimanual dexterous jar opening with two robot arms and dexterous hands.}
    \label{fig:teaser}
\end{figure}

Recent learning-based approaches to dexterous manipulation include both imitation learning (IL) and reinforcement learning (RL). IL methods learn dexterous and bimanual manipulation from human demonstrations~\cite{Zhao-RSS-23, ding2025bunny, li2025maniptrans}. However, collecting dexterous demonstrations is costly, and transferring human demonstrations to robotic hands introduces an embodiment gap due to differences in hand kinematics and morphology~\cite{pmlr-v305-lum25a, pmlr-v270-chen25f}. Current RL methods learn dexterous manipulation through large-scale interaction in simulation~\cite{andrychowicz2020learning}. RL has also been used for lid twisting task with two dexterous hands~\cite{linTwistingLidsTwo2025}. However, prior work considered a restricted setting where both robot arms were fixed and the object was placed directly on the fingers. By contrast, in our task setup shown in Fig.~\ref{fig:teaser}, the robot arms move the dexterous hands toward the jar, and the hands establish separate grasps on the body and lid before twisting begins.

We introduce Grasp2Twist, which learns a single policy through RL to jointly control both arms and hands from grasp acquisition through sustained lid twisting. The policy coordinates grasping and twisting.
Prior rewards based on fingertip proximity or finger contact~\cite{linTwistingLidsTwo2025,qin2023dexpoint} encourage grasp acquisition but do not explicitly account for the overall spatial arrangement of the palm and fingertips around the object.
Therefore, based on the spatial relationship between the object center and the convex hull formed by the hand, we introduce a continuous enclosure measure to guide this spatial arrangement, and a binary enclosure indicator to guide the grasp-to-twist transition.
We use curriculum learning with three stages. In the first two stages, we adjust rewards and termination conditions as learning progresses, first learning grasp acquisition and the transition to twisting, and then facilitating exploration of contact reconfiguration. In the final stage we progressively increase domain randomization to adapt the learned policy to real-world actuation and contact dynamics.

The learned policy transfers zero-shot to the physical robot and demonstrates finger gaiting through repeated cycles of twisting, release, repositioning, and recontact on a set of diverse real-world jars.
Ablation studies show that the enclosure measure and indicator improve grasp formation and that the curriculum makes contact-reconfiguration discovery more reliable.

Our main contributions are threefold:
\begin{itemize}
    \setlength{\itemsep}{1pt}
    \setlength{\parskip}{0pt}
    \setlength{\parsep}{0pt}
    \item A real-world bimanual dexterous jar-opening system with a unified policy operating from grasp acquisition until the lid threads disengage.
    \item A continuous enclosure measure and a binary enclosure indicator that guide grasp formation and the grasp-to-twist transition, respectively, within a unified policy.
    \item A three-stage curriculum that supports unified grasp-to-twist learning, contact-reconfiguration exploration, and robustness adaptation for sim-to-real deployment.
\end{itemize}

\section{Related Work}
\label{sec:related_work}

\textbf{Bimanual Dexterous Manipulation.}
Bimanual dexterity increases control dimensionality and introduces inter-hand coordination, particularly when both hands manipulate a shared object ~\cite{krebs2022bimanual}. IL methods learn manipulation policies from demonstrations collected through teleoperation, motion capture, and human video~\cite{Zhao-RSS-23,li2025maniptrans,ding2025bunny,qin2023anyteleop,zheng2026egoscale}. RL methods learn manipulation policies through interaction with simulated environments and have demonstrated single-hand reorientation~\cite{andrychowicz2020learning,chen2023visual,Khandate-RSS-23}, tactile-based in-hand rotation~\cite{Yin-RSS-23}, and bimanual lid twisting~\cite{linTwistingLidsTwo2025}.

\textbf{Dexterous Grasp.}
Classical grasp analysis assesses grasp quality from contact geometry through force-closure and wrench-space criteria~\cite{ferrari1992planning}, while differentiable force closure enables both grasp synthesis and grasp-quality evaluation~\cite{liu2021synthesizing,li2023gendexgrasp,zhai20222}. Geometric caging characterizes whether finger configurations prevent object escape~\cite{makita20083d}. Learning-based approaches to dexterous grasping use contact-based rewards to encourage coordinated finger contact~\cite{qin2023dexpoint} and geometric interaction representations to capture finger--object spatial relations~\cite{she2022learning}. Task-oriented grasping additionally considers whether a grasp supports subsequent manipulation rather than immediate stability alone~\cite{fang2020learning}. Our method uses a continuous enclosure measure to guide grasp formation and a binary enclosure indicator to define the transition to downstream manipulation.

\textbf{Curriculum Learning in Robot Learning.}
Curriculum learning facilitates policy optimization by structuring training from easier to more difficult conditions~\cite{bengio2009curriculum}. In reinforcement learning for robotics, curriculum strategies include progressively expanding initial state distributions~\cite{florensa2017reverse}, increasing environmental difficulty such as increasing terrain complexity~\cite{rudin2022learning}, broadening dynamics randomization for sim-to-real transfer~\cite{akkaya2019solving}, and introducing physical difficulty to facilitate dexterous manipulation~\cite{chen2022system}. Our curriculum combines task progression and robustness adaptation.

\section{Task and System Overview}
\label{sec:system}

\begin{figure*}[t]
    \centering
    \includegraphics[width=\textwidth]{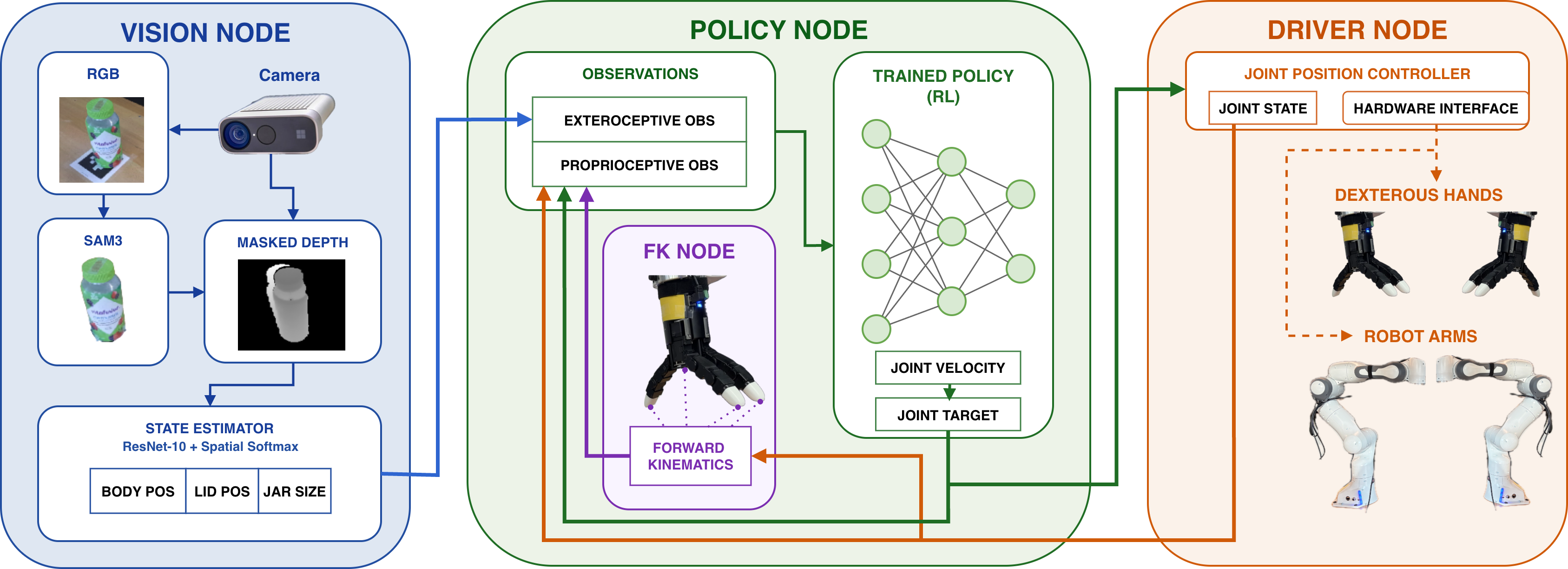}
    \caption{Overview of the real-world system. The vision node estimates jar-body and lid poses and dimensions from segmented depth images. The policy node combines these estimates with joint states, joint position targets, and end-effector poses to generate actions. The driver node tracks the resulting joint position targets and returns joint states for policy observations and forward kinematics.
    }
    \label{fig:system_diagram}
\end{figure*}

This section formulates the bimanual jar-opening task and describes the real-world system and simulation environment. It then discusses the challenges in policy learning and sim-to-real transfer that motivate the proposed method in this paper.

\subsection{Task Formulation}
\label{subsec:task}
Starting from pre-grasp configurations, the two arms move their mounted hands toward the jar and establish separate grasps on the jar body and lid. The body-side hand, referred to as the \textit{holding hand}, stabilizes the jar body, while the lid-side hand, referred to as the \textit{twisting hand}, rotates the lid in the opening direction. Success in the jar-opening task is defined as complete lid thread disengagement from the jar body while the jar body remains held by the \textit{holding hand}.

The task is formulated as a partially observable Markov decision process (POMDP). At each policy step $t$, the policy $\pi_{\theta}$ receives proprioceptive observations $\mathbf{o}_{\mathrm{propri},t}$ and exteroceptive observations $\mathbf{o}_{\mathrm{extero},t}$, and outputs normalized actions $\mathbf{a}_{t}$ in a high-dimensional action space $\mathcal{A} \coloneqq [-1, 1]^{38}$ for the actuated joints from both arms and hands. The actions are smoothed using an EMA coefficient of $0.5$, rescaled using the policy timestep and predefined velocity and action scales to obtain joint position increments $\Delta \mathbf{q}_{t+1}$, and converted into joint position targets through leaky integration:
\begin{equation}
    \mathbf{q}_{\mathrm{target}, t+1} = \alpha \, \mathbf{q}_{\mathrm{target}, t} 
    + (1 - \alpha) \, \mathbf{q}_t 
    + \Delta \mathbf{q}_{t+1}
\label{eq:action_integration}
\end{equation}
where $\mathbf{q}_t$ denotes the current joint positions, $\mathbf{q}_{\mathrm{target},t}$ denotes the joint position targets at step $t$, and $\alpha\in[0,1]$ is the leaky-integration factor that interpolates between $\mathbf{q}_{\mathrm{target},t}$ and $\mathbf{q}_t$.

\subsection{Real-World System}
\label{subsec:hardware}

\textbf{Hardware.} 
As illustrated in Fig.~\ref{fig:system_diagram}, the hardware platform consists of two Franka Emika Panda arms mounted on opposite sides of a shared tabletop workspace. Each arm is equipped with a TESOLLO DG-3F-B three-fingered hand. The trained policy updates joint position targets at $20\,\mathrm{Hz}$, while the low-level arm and hand controllers track these targets at $1000\,\mathrm{Hz}$ and $200\,\mathrm{Hz}$, respectively. The end-effector poses are computed from forward kinematics at $100\,\mathrm{Hz}$. At each policy step, the latest joint states, joint position targets, and end-effector poses are collected together. The most recent eight steps of these quantities constitute the proprioceptive observations $\mathbf{o}_{\mathrm{propri},t}$ used by the policy. 

\textbf{Perception.}
A depth-based perception pipeline provides object-centric observations for real-world deployment. An Azure Kinect camera captures RGB-D observations, from which SAM3 with TensorRT segments the jar body and lid~\cite{carion2026sam}. The masked depth observations are processed by a ResNet-10-based regressor with spatial softmax to estimate the poses and dimensions of the jar body and lid~\cite{he2016deep, levine2016end}. These estimates constitute the exteroceptive observations $\mathbf{o}_{\mathrm{extero},t}$ used by the policy.

\subsection{Simulation Environment}
\label{subsec:simulation}

The simulation environment reconstructs the bimanual robot platform using URDF and MJCF models in mjlab~\cite{zakka2026mjlablightweightframeworkgpuaccelerated} and MuJoCo~\cite{todorov2012mujoco}. Each jar is modeled using two coaxial cylinders corresponding to the jar body and lid~\cite{linTwistingLidsTwo2025}. The lid is connected to the jar body by a revolute joint. The threads are not explicitly modeled in simulation. Joint damping is used to approximate the rotational resistance. The simulation randomizes jar geometry and physical properties. The state estimator in
Sec.~\ref{subsec:hardware} is trained on masked depth images rendered in
simulation, using randomized jar geometries and poses as supervision.

\subsection{Challenges}
\label{subsec:challenges}

Learning bimanual dexterous jar opening raises three challenges: unified grasp-to-twist learning, contact-reconfiguration exploration, and sim-to-real transfer.

\subsubsection{\textbf{Unified Grasp-to-twist Learning}}
\label{subsubsec:challenges_transition}
The system uses a single policy for grasp acquisition and lid twisting without an explicit task-phase input.
Grasp acquisition may end in a wide range of hand and object configurations~\cite{chen2023sequential}, especially in high-dimensional dexterous manipulation. The policy therefore needs to learn an appropriate transition from grasping to twisting within the unified policy.

\subsubsection{\textbf{Contact-Reconfiguration Exploration}} 
\label{subsubsec:challenges_reconfiguration} 
Maintaining a fixed finger-contact configuration eventually exhausts the feasible motion range of the \textit{twisting hand} under kinematic constraints, making contact reconfiguration necessary for sustained lid twisting. Such reconfiguration can involve releasing one or more fingers, repositioning them, and re-establishing contact before twisting continues. Learning these contact changes poses a difficult exploration problem for RL~\cite{Khandate-RSS-23}. In our task, breaking contacts that support grasp acquisition can temporarily reduce the reward in RL, while the benefit of reconfiguring contact appears only after new contacts are established and twisting resumes.

\subsubsection{\textbf{Sim-to-Real Transfer}}
\label{subsubsec:challenges_sim2real}
Transferring the unified policy to the physical system requires preserving effective contact reconfiguration and bimanual stability despite discrepancies in actuation and contact dynamics. Contact reconfiguration emerges during learning, and the current learning objective cannot guarantee sufficient separation between the fingers and the lid when releasing contacts. On hardware, tracking errors and contact-model mismatch may leave insufficient clearance for the fingers to break contacts and reposition. Additionally, interactions between the twisting fingers and the lid generate disturbances that propagate through the shared jar to the \textit{holding hand}. Compared with simulation, these disturbances can induce larger jar motion in the real world and destabilize the holding grasp.

\section{Method}
\label{sec:rl_method}

\begin{figure*}[t]
    \centering
    \includegraphics[width=\textwidth]{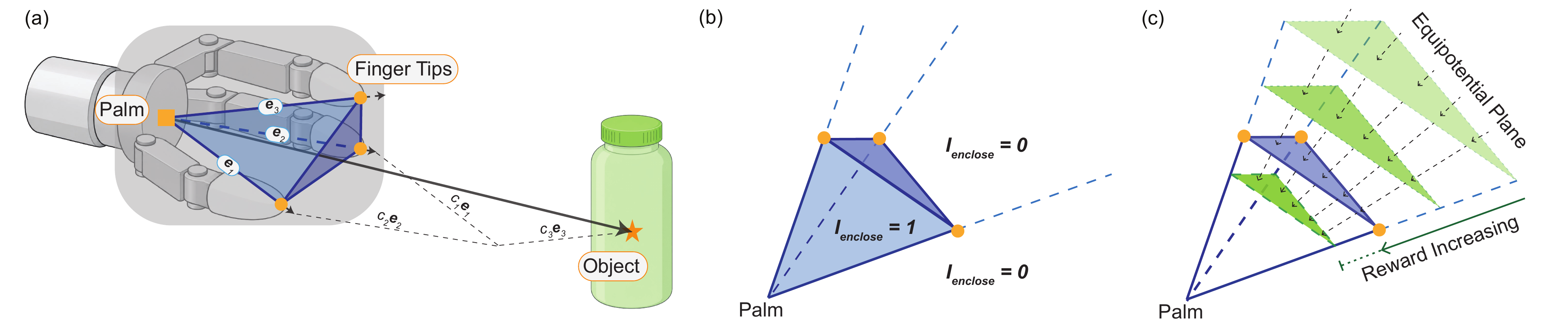}
    \caption{Geometric hand--object enclosure representation.
    (a) The palm position $\mathbf{p}_{\mathrm{palm}}$ and three fingertip positions $\mathbf{p}_{\mathrm{tip},i}$ define a tetrahedral convex hull;
    the displacement from the palm $\mathbf{p}_{\mathrm{palm}}$ to the object center $\mathbf{p}_{\mathrm{obj}}$ is decomposed along the three palm-to-fingertip edge vectors $\mathbf{e}_i$.
    (b) The binary enclosure indicator $\mathbb{I}_{\mathrm{enclose}}$ identifies whether the object center lies inside the tetrahedron convex hull.
    (c) The geometric enclosure reward derived from the enclosure measure $m_{\mathrm{enclose}}$.
    }
    \label{fig:geom_representation}
\end{figure*}

The policy is trained using PPO with an asymmetric actor--critic architecture through RSL-RL~\cite{schulman2017proximal, pinto2017asymmetric,schwarke2025rslrl}. The actor and critic are three-layer MLPs with hidden dimensions $[512,256,128]$. Training uses 4096 parallel environments with 128 environment steps per PPO iteration. The actor receives noisy $\mathbf{o}_{\mathrm{propri},t}$ and $\mathbf{o}_{\mathrm{extero},t}$, while the critic accesses noise-free observations with additional privileged states during training.

\subsection{Geometric Hand--Object Enclosure Representation}
\label{subsec:geometric_representation}

The geometric hand--object enclosure representation describes the geometric enclosure state in which the object center lies within the convex hull formed by the hand, with a binary indicator and a continuous measure. The indicator determines whether the grasp satisfies the enclosure condition, while the measure quantifies the progress toward this condition.

\subsubsection{\textbf{Convex Hull Enclosure Indicator}}
Each three-fingered hand is represented by the tetrahedral convex hull defined by the palm position $\mathbf{p}_{\mathrm{palm}}$ and the three fingertip positions $\mathbf{p}_{\mathrm{tip},i}$, $i\in\{1,2,3\}$, as illustrated in Fig.~\ref{fig:geom_representation}(a). Let $\mathbf{e}_i=\mathbf{p}_{\mathrm{tip},i}-\mathbf{p}_{\mathrm{palm}}$ denote the three palm-to-fingertip edge vectors. The displacement from the palm to the object center $\mathbf{p}_{\mathrm{obj}}$ is decomposed along those three edge vectors $\mathbf{e}_i$ as
\begin{equation}
    \mathbf{p}_{\mathrm{obj}}-\mathbf{p}_{\mathrm{palm}}
    =
    \sum_{i=1}^{3} c_i\mathbf{e}_i.
    \label{eq:enclosure_decomposition}
\end{equation}
The coefficients $c_i$ specify the position of the object center with
respect to this geometric basis.

The object center lies inside the tetrahedral convex hull if $c_i\geq0$ for all $i$ and $\sum_{i=1}^{3}c_i\leq1$. Accordingly, the binary enclosure indicator is defined as

\begin{equation}
    \mathbb{I}_{\mathrm{enclose}}
    =
    \mathbb{I}\!\left[
        \forall i\in\{1,2,3\},\ c_i \geq 0,
        \quad
        \sum_{i=1}^{3} c_i \leq 1
    \right], 
    \label{eq:enclosure_indicator}
\end{equation}
where $\mathbb{I}[\cdot]$ denotes the indicator function. Fig.~\ref{fig:geom_representation}(b) illustrates the enclosed and non-enclosed cases.

Rather than explicitly evaluating contact wrenches~\cite{ferrari1992planning}, the proposed indicator provides a geometric proxy for whether the hand surrounds the object and serves as one transition condition from grasp acquisition to lid twisting. We use 
$\mathbb{I}_{\mathrm{enclose}}^{\mathrm{hold}}$ and $\mathbb{I}_{\mathrm{enclose}}^{\mathrm{twist}}$ to indicate whether the \textit{holding hand} encloses the jar body and the \textit{twisting hand} encloses the lid, respectively.

\subsubsection{\textbf{Enclosure Measure}}
When $c_i\geq0$ for all $i$, the coefficient sum defines a continuous enclosure measure,
\begin{equation}
    m_{\mathrm{enclose}}
    =
    \sum_{i=1}^{3} c_i.
    \label{eq:enclosure_measure}
\end{equation}
Geometrically, the measure is the ratio of the palm's distances to two parallel planes: one through the object center and the other through the three fingertips. The set $m_{\mathrm{enclose}}=1$ corresponds to the fingertip face, $0\leq m_{\mathrm{enclose}}<1$ to points inside the tetrahedron, and
$m_{\mathrm{enclose}}>1$ to points beyond the fingertip face.

During grasp acquisition, decreasing $m_{\mathrm{enclose}}$ from values above one to below one therefore guides the object center across the fingertip face and into the enclosure region.

\subsection{Reward Design}
\label{subsec:reward}

The reward is designed around grasp acquisition and lid twisting. The enclosure measure and finger contact rewards guide grasp acquisition, while lid angular velocity rewards twisting progress. The geometric enclosure indicator gates the twisting reward to encode the grasp-to-twist transition.

\subsubsection{\textbf{Grasp Acquisition}}
We adapt the finger contact reward from prior work~\cite{linTwistingLidsTwo2025} to our task using a clipped exponential decay function:
\begin{equation}
    r_{\mathrm{cont}} =
    \exp \left(-\frac{\operatorname{clip}\left(\bar d,\, d_{\min},\, d_{\max}\right)-d_{\min}}{\sigma}
    \right),
    \label{eq:contact_reward}
\end{equation}
where $\bar d$ is the mean distance from fingertips to their nearest surface sample points, $d_{\min}$ and $d_{\max}$ are clipping thresholds, and $\sigma$ is the decay length.

However, the surface proximity does not explicitly encourage the fingers to form a spatial arrangement around the object. As illustrated in Fig.~\ref{fig:geom_representation}(c), we introduce a geometric enclosure reward derived from Eq.~\eqref{eq:enclosure_measure}:
\begin{equation}
    r_{\mathrm{enclose}} = \mathbb{I}_{\{c_i \ge 0\}} \cdot
    \exp \left(-\frac{\operatorname{clip}\left(m_{\mathrm{enclose}},\, m_{0},\, m_{\max}\right)-m_{0}}{\tau}
    \right),
    \label{eq:enclosure_reward}
\end{equation}
where $m_{0}$ defines the reward plateau threshold and lower clipping bound, $m_{\max}$ is the upper clipping bound, and $\tau$ controls the decay scale.

\subsubsection{\textbf{Lid Twisting}}

Twisting progress is measured by the lid angular velocity $\omega_{\mathrm{lid}}$, with the opening direction defined as positive. 
We obtain the clipped and normalized angular velocity $\tilde{\omega}_{\mathrm{lid}}$ as
\begin{equation}
    \tilde{\omega}_{\mathrm{lid}}
    =
    \frac{
        \operatorname{clip}
        \left(
            \omega_{\mathrm{lid}},
            -\omega_{\mathrm{tol}},
            \omega_{\mathrm{max}}
        \right)
    }{
        \omega_{\mathrm{max}}
    },
    \label{eq:lid_velocity_reward}
\end{equation}
where $-\omega_{\mathrm{tol}}=-0.05\pi\,\mathrm{rad/s}$ limits the penalty for transient backward rotation during finger exploration, and $\omega_{\mathrm{max}}=0.5\pi\,\mathrm{rad/s}$ caps the rewarded opening velocity.

To preserve the grasp-to-twist ordering, the twisting reward is activated only when the \textit{holding hand} encloses the jar body, the \textit{twisting hand} encloses the lid, and the contact count $N_{\mathrm{contact}}$ between the fingers on \textit{twisting hand} and the lid is at least two:
\begin{equation}
    r_{\mathrm{task}}
    =
    \mathbb{I}_{\mathrm{enclose}}^{\mathrm{hold}}
    \cdot
    \mathbb{I}_{\mathrm{enclose}}^{\mathrm{twist}}
    \cdot
    \mathbb{I}_{\{N_{\mathrm{contact}}\geq 2\}}
    \cdot
    \tilde{\omega}_{\mathrm{lid}}.
    \label{eq:twisting_reward}
\end{equation}

The overall reward combines the task, enclosure, finger contact, and regularization terms: 
\begin{equation}
    R_{\mathrm{total}} = R_{\mathrm{task}} + R_{\mathrm{enclose}} + R_{\mathrm{cont}} + R_{\mathrm{reg}}, 
\end{equation} 
where superscripts $\mathrm{hold}$ and $\mathrm{twist}$ denote the \textit{holding hand} interacting with the jar body and the \textit{twisting hand} interacting with the lid, respectively, $R_{\mathrm{task}}=12\,r_{\mathrm{task}}$, $R_{\mathrm{enclose}}=2\,r_{\mathrm{enclose}}^{\mathrm{hold}}+r_{\mathrm{enclose}}^{\mathrm{twist}}$, and $R_{\mathrm{cont}}=r_{\mathrm{cont}}^{\mathrm{hold}}+0.625\,r_{\mathrm{cont}}^{\mathrm{twist}}$; $R_{\mathrm{reg}}$ sums the weighted regularization terms.

\subsection{Curriculum Learning}
\label{subsec:curriculum}

Training follows a three-stage curriculum for foundational skill acquisition,
contact-reconfiguration exploration, and sim-to-real adaptation.

\subsubsection{\textbf{Foundational Skill Acquisition}}
Curriculum Stage~1 learns grasp acquisition, the grasp-to-twist transition, and initial lid twisting. It relaxes actuator effort, motion regularization, and termination conditions to facilitate initial exploration. Most sim-to-real randomization is deferred to Stage~3.

\subsubsection{\textbf{Contact-reconfiguration Exploration}}
Stage~2 begins at iteration 2500 and facilitates contact-reconfiguration exploration by widening the saturation threshold $d_{\min}$ of $r_{\mathrm{cont}}^{\mathrm{twist}}$ from $0.005\,$m to $0.025\,$m, reducing the reward loss associated with temporary finger--object separation. We also increase motion regularization for the arms and the \textit{holding hand} and tighten constraints on jar motion. Regularization on the twisting fingers remains low, concentrating exploration on changes in their contacts while preserving stability of the rest of the bimanual system.

\subsubsection{\textbf{Sim-to-Real Adaptation}}
Stage~3 begins at iteration 5000 and adapts the policy for sim-to-real transfer. After the contact reconfiguration has been learned, we progressively broaden randomization over actuator effort, controller gains, gravity compensation, joint and contact dynamics, and observation noises, while increasing regularization and tightening termination
conditions. This gradual transition avoids sudden shifts in the training settings. The final randomization ranges are summarized in Table~\ref{tab:domain_randomization}.

\begin{table}[t]
    \centering
    \caption{Final domain randomization ranges.}
    \label{tab:domain_randomization}
    \small
    \setlength{\tabcolsep}{3pt}
    \renewcommand{\arraystretch}{1.03}
    \begin{tabular}{@{}l c c@{}}
        \hline
        \textbf{Parameter} & \textbf{Mode} & \textbf{Final range} \\
        \hline

        Jar-body mass (kg)
            & Value & $[0.05,\ 0.30]$ \\
        Jar friction
            & Value & $[0.20,\ 2.50]$ \\
        Lid-joint damping
            & Value & $[0.10,\ 0.50]$ \\
        Jar body radius (m)
            & Value & $[0.03,\ 0.05]$ \\
        Jar body height (m)
            & Value & $[0.10,\ 0.20]$ \\
        Lid radius (m)
            & Value & $[0.02,\ 0.06]$ \\
        Lid height (m)
            & Value & $[0.02,\ 0.04]$ \\
        \hline

        Action noise
            & Add. & $\pm0.10$ \\
        Action delay (steps)
            & Value & $[0,\ 2]$ \\
        Finger actuator effort
            & Value & $[0.82,\ 1.02]$ \\
        Arm/finger PD gains $(K_p,K_d)$
            & Scale & $[0.75,\ 1.10]$ \\
        Arm joint friction loss
            & Value & $[0.05,\ 0.50]$ \\
        Finger joint friction loss
            & Value & $[0.05,\ 0.15]$ \\
        Gravity compensation
            & Scale & $[0.90,\ 1.00]$ \\
        \hline

        Joint position noise
            & Add. & $\pm0.0025\;/\;\pm0.020$ \\
        FK position noise (m)
            & Add. & $\pm0.001\;/\;\pm0.015$ \\
        Object position noise (m)
            & Add. & $\pm0.010\;/\;\pm0.025$ \\
        Jar geometry noise (normalized)
            & Add. & $\pm0.005\;/\;\pm0.200$ \\
        \hline

        \multicolumn{3}{@{}l@{}}{\footnotesize
        Observation noise ranges are reported as uniform noise / additive bias.}
    \end{tabular}
\end{table}

\section{Experiments}
\label{sec:experiments}

This section evaluates the proposed system through real-world deployment and ablation studies. The real-world experiments assess the task performance across diverse jar properties. The ablation studies evaluate the contributions of the geometric enclosure representation (Sec.~\ref{subsec:geometric_representation}) and the curriculum learning (Sec.~\ref{subsec:curriculum}).

\subsection{Real-World Evaluation}
We evaluate the learned policy on the physical system using both in-distribution and geometrically out-of-distribution objects. 

\subsubsection{\textbf{Experimental Setup and Protocol}}
\begin{figure}[t]
    \centering
    \includegraphics[width=\columnwidth]{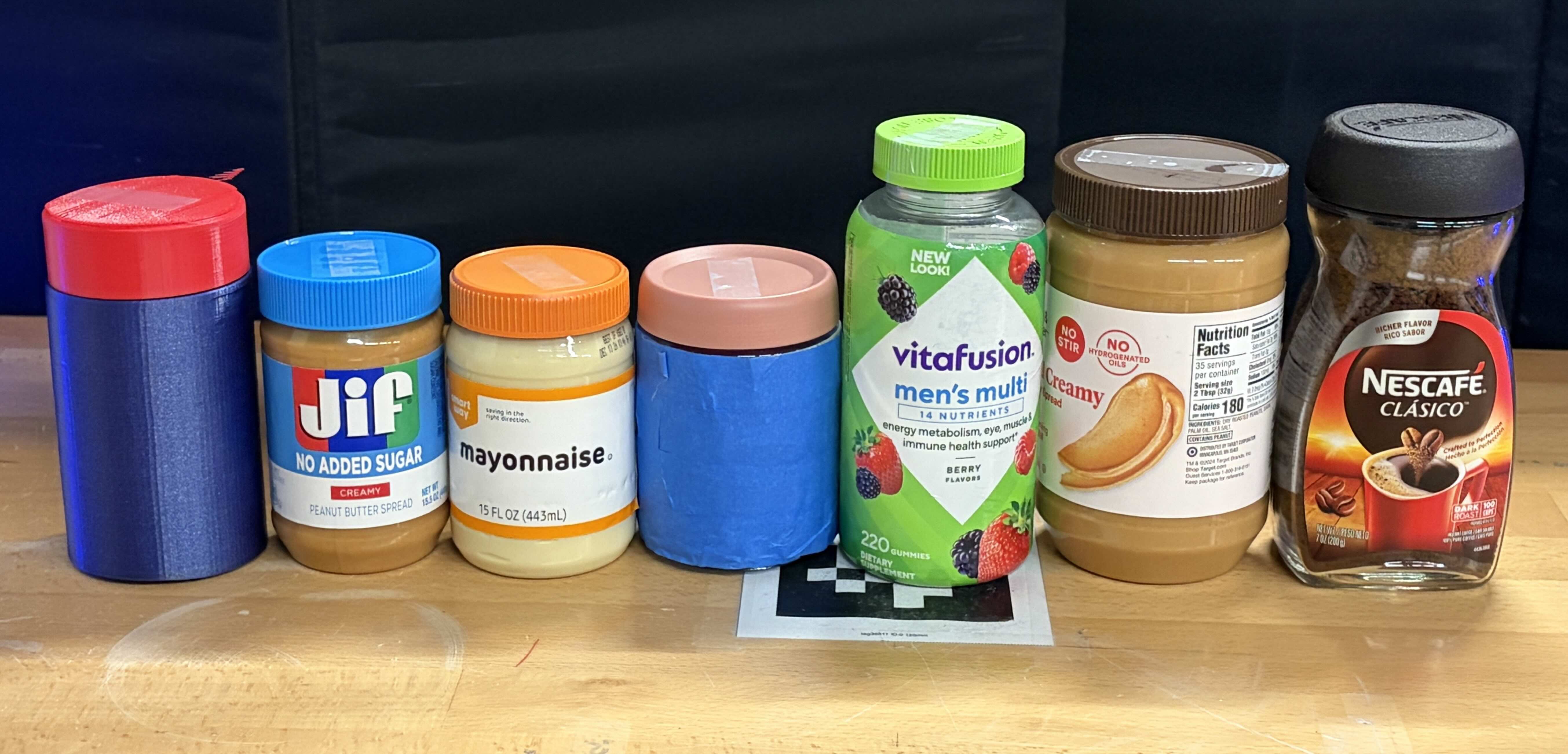}
    \caption{
    Real-world test instances. The leftmost object is the ID 3D-printed jar. The remaining objects are OOD-Geom household containers, referred to as Blue Peanut, Yellow Mayo, Pink Glass, Green Vitamin, Brown Peanut, and Instant Coffee.
    }
    \label{fig:real_world_instances}
\end{figure}

The experiments are conducted on the physical system described in Sec.~\ref{subsec:hardware}. All real-world experiments use a single policy trained with random seed $42$. The same checkpoint is evaluated on one 3D-printed jar and six off-the-shelf household containers shown in Fig.~\ref{fig:real_world_instances}. The 3D-printed jar matches the cylindrical object model used in simulation and constitutes the in-distribution (ID) test set. The household containers differ in shape from the simplified cylindrical model and constitute the geometrically out-of-distribution (OOD-Geom) test set.

\begin{figure*}[t]
    \centering
    \includegraphics[width=\textwidth]{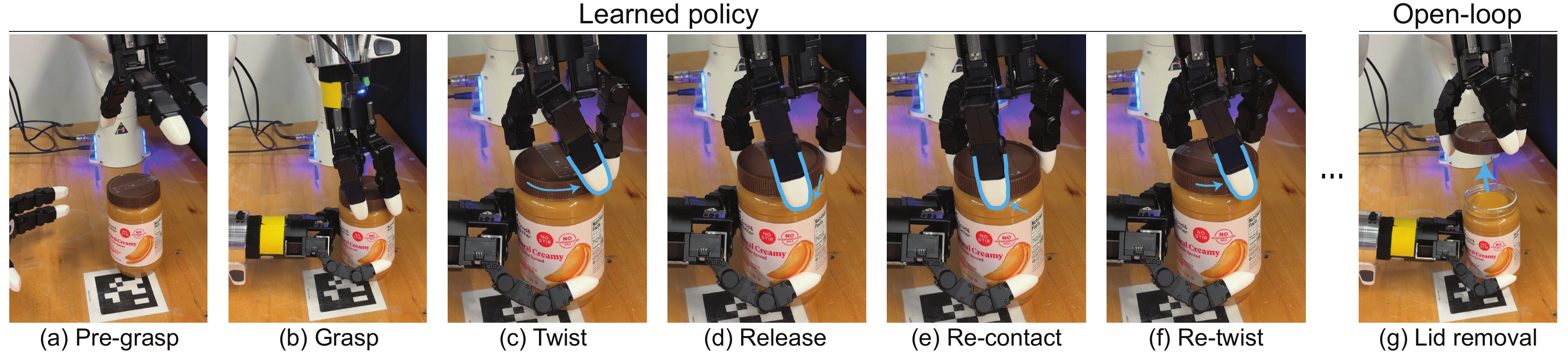}
    \caption{
    Real-world execution sequence from pre-grasp to lid removal. After grasp acquisition, the \textit{holding hand} stabilizes the jar body while the \textit{twisting hand} rotates the lid. Individual fingers asynchronously twist, release, reposition, and re-establish contact. Twisting continues until complete thread disengagement, after which the lid is removed open loop.
    }
    \label{fig:real_sequence}
\end{figure*}

\begin{table*}[t]
    \centering
    \caption{Real-World Evaluation on OOD-Geom Household Containers}
    \label{tab:real_world_ood}
    \resizebox{\textwidth}{!}{
        \begin{tabular}{lcccccc}
            \hline
            \textbf{Metric}
            & \textbf{Blue Peanut}
            & \textbf{Yellow Mayo}
            & \textbf{Pink Glass}
            & \textbf{Green Vitamin}
            & \textbf{Brown Peanut}
            & \textbf{Instant Coffee} \\
            \hline\hline

            Approx. Rotation to Disengagement
            & $2\pi$ & $4\pi$ & $\pi$ & $1.5\pi$ & $4\pi$ & $\pi$ \\

            Jar-Opening Success (Replay)
            & 1/10 & 2/10 & 0/10 & 3/10
            & 6/10 & 6/10 \\

            Jar-Opening Success (Ours)
            & 8/10 & 9/10 & 6/10 & 10/10
            & 10/10 & 10/10 \\

            Lid Angular Velocity (Ours, Mean $\pm$ Std)
            & $(0.19 \pm 0.01)\pi$/s & $(0.26 \pm 0.02)\pi$/s & $(0.06 \pm 0.01)\pi$/s & $(0.22 \pm 0.02)\pi$/s & $(0.19 \pm 0.05)\pi$/s & $(0.21 \pm 0.01)\pi$/s \\

        \end{tabular}
    }
\end{table*}

Each instance is evaluated over ten trials, with its initial position randomized within a $0.2\,\mathrm{m}\times0.2\,\mathrm{m}$ tabletop region. The robot starts each trial from the same predefined pre-grasp configuration.

The \textbf{ID evaluation} measures sustained twisting on a 3D-printed jar. The lid remains mechanically engaged throughout each $30\,\mathrm{s}$ trial, allowing continuous rotation without termination by thread disengagement. We record the cumulative lid rotation and the number of finger-gait events over the evaluation horizon. A finger-gait event is counted when a finger of the \textit{twisting hand} releases the lid, repositions, and re-establishes contact, followed by continued opening rotation from the new contact configuration.

The \textbf{OOD-Geom evaluation} measures jar-opening success on household containers. Before each trial, the lid is manually rotated in the closing direction under light manual force until no further rotation is observed; no locking mechanism or additional tightening force is applied.

Each trial terminates upon complete thread disengagement, a grasp failure that prevents further twisting, or a time limit of $30\,\mathrm{s}$. A trial is considered successful when the lid can subsequently be removed from the jar body by the \textit{twisting hand} without additional rotation or observable resistance, while the jar body remains held by the \textit{holding hand}. The lid removal is performed open loop.

For each instance, we report the jar-opening success rate and the mean lid
angular velocity over successful trials. We additionally report the approximate lid rotation from the initial threaded state to thread disengagement, obtained by manually rotating the lid until it lifts free and rounding to the nearest $\pi/2$. This quantity characterizes the thread-engagement span and is not used as the success criterion.

To assess the contribution of closed-loop policy feedback, we additionally evaluate a non-reactive replay baseline. The baseline replays on the physical system the joint target trajectories that were successful in simulation. Prior lid-twisting work~\cite{linTwistingLidsTwo2025} assumes fixed arms and object initialized on fingers, making its reported results not directly comparable to our evaluation. Our contact-reward-only baseline, only using the finger contact reward $r_{\mathrm{cont}}$ adapted from that work, failed to acquire jar-body grasps in our simulation experiments (Sec.~\ref{subsubsec:ablation_enclosure}), so we did not deploy it on the physical system.

\subsubsection{\textbf{Results}}
Across ten trials on the \textbf{ID} 3D-printed jar, the policy achieves a mean cumulative lid rotation of $(7.13 \pm 0.21)\pi$ rad and $45.5 \pm 2.66$ finger-gait events over the $30\,\mathrm{s}$ evaluation horizon. These results demonstrate sustained twisting and repeated finger gaiting on the physical system for the ID geometry.

Table~\ref{tab:real_world_ood} reports the results on the six \textbf{OOD-Geom} household containers. The learned policy succeeds in 53/60 trials (88\%), compared with 18/60 trials (30\%) for the replay baseline. Both methods achieve lower success rates on relatively smaller containers---Blue Peanut, Yellow Mayo, and Pink Glass---with a substantially larger degradation under replay. The learned policy achieves per-instance success rates from 6/10 to 10/10 across all six objects.

Fig.~\ref{fig:real_sequence} illustrates the physical execution of the learned policy, from grasp acquisition, through detailed behaviors during sustained lid twisting, to open-loop lid removal.

Table~\ref{tab:state_estimation} summarizes static estimation errors on seven instances, each evaluated with and without dexterous-hand occlusion, using manually measured values as ground truth. Occlusion generally increases the mean error in vertical position and object dimensions, while its effect on the overall 3D position error varies between ID and OOD-Geom objects. Across all objects, we also observe a systematic positional bias of approximately $20.5\,\mathrm{mm}$.

\begin{table}[h]
    \centering
    \caption{Static State Estimation Errors (mean $\pm$ std).}
    \label{tab:state_estimation}
    \small
    \setlength{\tabcolsep}{4pt}
    \begin{tabular}{lcc}
        \hline
        \textbf{Error}
        & \textbf{Unoccluded(mm)}
        & \textbf{Occluded(mm)} \\
        \hline\hline
        ID 3D position
        & $24.2 \pm 10.3$ & $22.9 \pm 6.4$ \\
        OOD-Geom 3D position
        & $26.9 \pm 10.7$ & $29.3 \pm 12.5$ \\
        \hline
        ID vertical position
        & $1.9 \pm 0.9$ & $3.3 \pm 3.0$ \\
        OOD-Geom vertical position
        & $7.5 \pm 6.0$ & $14.6 \pm 11.9$ \\
        ID dimensions
        & $4.3 \pm 2.6$ & $5.5 \pm 3.4$ \\
        \hline
    \end{tabular}
\end{table}

\begin{figure*}[t]
    \centering
    \includegraphics[width=\textwidth]{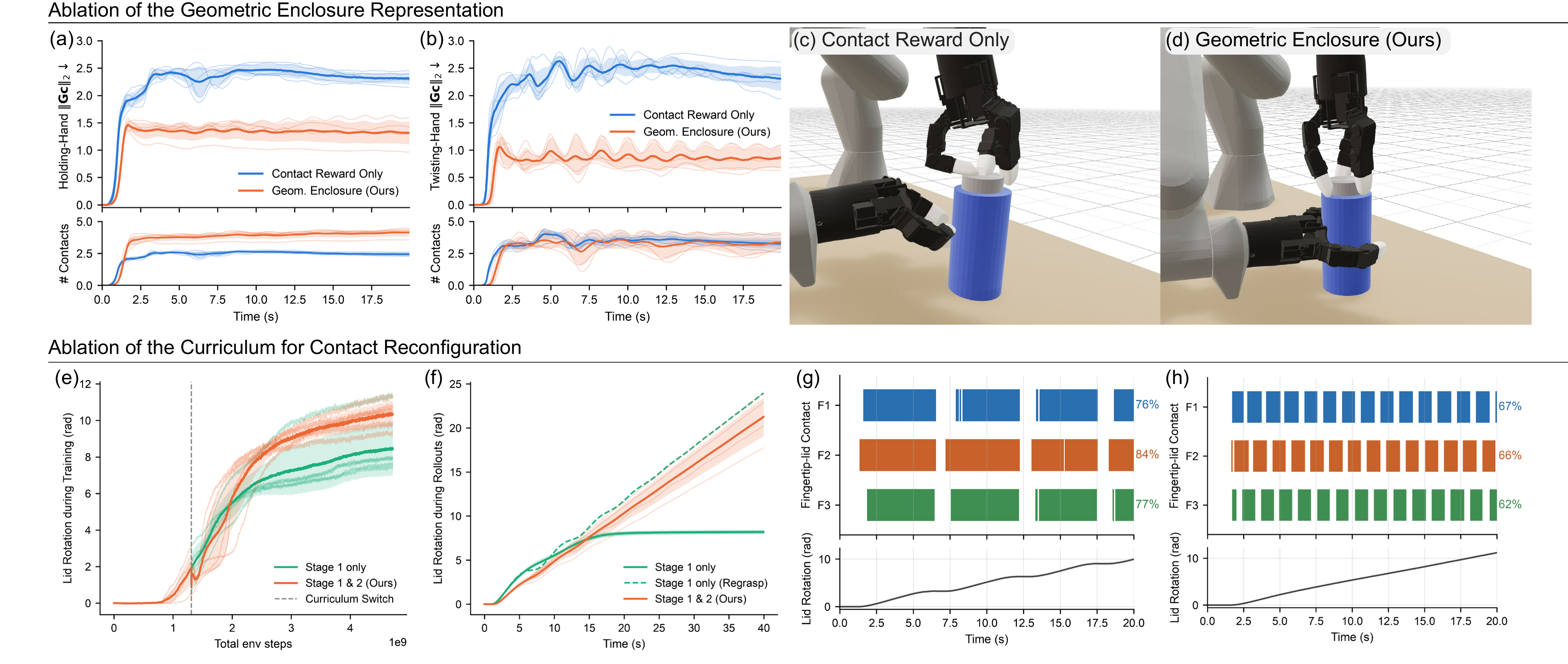}
    \caption{
    Ablation studies of the geometric enclosure representation and the curriculum for contact reconfiguration.
    (a)--(b) DFC residuals $\|\mathbf{G}\mathbf{c}\|_2$ of the \textit{holding hand} and \textit{twisting hand} during evaluation rollouts for the contact-reward-only baseline and full method with geometric enclosure; the lower subpanels show the corresponding number of hand--object contacts.
    (c)--(d) Representative grasps from the contact-reward-only baseline and the full method, respectively, showing the different spatial arrangements of the fingers around the object.
    (e) Lid rotation during training under two curriculum settings, Stage~1 only and Stages~1 and~2.
    (f) Lid rotation during evaluation rollouts for policies trained with Stage~1 only and with Stages~1 and~2. The dashed curve shows the single Stage-1-only seed that discovers contact reconfiguration, separated from the remaining four seeds.
    (g)--(h) Representative contact-reconfiguration behaviors observed in simulation: synchronous contact changes in twist-and-regrasp behavior and asynchronous contact changes in finger gaiting; percentages give each finger's contact time fraction.
    Curves show the mean over seeds; shaded regions denote one standard deviation across seeds; each seed's curve is itself averaged over 4096 parallel environments.
    }
    \label{fig:ab_study}
\end{figure*}

We inspect the unsuccessful policy trials using the recorded videos. Most failures occur on smaller containers. Observed failures include cases where the \textit{twisting hand} approaches the lid at an incorrect position or fails to establish a stable lid grasp. Segmentation degradation is visible in some trials, and state estimation errors may contribute to some of these grasp acquisition failures. We also observe the \textit{twisting hand} perturbing the jar, followed by loss of the \textit{holding hand} grasp. Similar grasp acquisition failures and grasp loss are observed in the replay trials.

\subsection{Ablation Study}
The ablation studies evaluate the geometric enclosure representation and the curriculum for contact reconfiguration introduced in Sec.~\ref{sec:rl_method}. We conduct both ablation studies in simulation using $4096$ parallel environments and five random seeds per training condition.

\subsubsection{\textbf{Ablation of the Geometric Enclosure Representation}}
\label{subsubsec:ablation_enclosure}
We evaluate the effect of the geometric enclosure representation by comparing the full method with a contact-reward-only baseline. The contact-reward-only baseline retains the finger contact reward $r_\mathrm{cont}$ adapted from prior lid-twisting work~\cite{linTwistingLidsTwo2025}, but removes the geometric enclosure reward $r_{\mathrm{enclose}}$ and binary enclosure indicator $\mathbb{I}_{\mathrm{enclose}}$. All other training settings are kept unchanged.

The resulting grasp configurations are assessed using the differentiable force-closure (DFC) residual, $r_{\mathrm{DFC}}=\|\mathbf{G}\mathbf{c}\|_2$, where $\mathbf{G}$ is the grasp matrix and $\mathbf{c}$ contains the unit surface normals at the contact points ~\cite{liu2021synthesizing}. Lower values indicate lower wrench imbalance under the DFC approximation.

Figs.~\ref{fig:ab_study}(a) and (b) report the rollout DFC residuals together with the corresponding numbers of hand--object contacts for the \textit{holding hand} and \textit{twisting hand}. Compared to the baseline, the proposed method achieves lower DFC residuals for both hands, indicating more balanced contact-normal wrenches.

Figs.~\ref{fig:ab_study}(c) and (d) illustrate that the baseline policy can satisfy local fingertip contact objectives by pressing against the object surface, whereas the proposed method produces enclosing grasps around the jar body and lid.

\subsubsection{\textbf{Ablation of Curriculum for Contact Reconfiguration}}
We evaluate the effect of curriculum Stage~2, which is designed to facilitate exploration of contact reconfiguration, by comparing policies trained only under Stage~1 with policies trained under Stage~1 followed by Stage~2. The former retains the Stage~1 objective throughout training, while the latter transitions to the Stage~2 objective after $2500$ training iterations. Neither setting proceeds to Stage~3; both use the same total number of environment steps and otherwise identical training settings.

Figs.~\ref{fig:ab_study}(e) and (f) show lid rotation during training and evaluation rollouts, respectively. Among the policies trained only under Stage~1, one of five seeds discovers contact reconfiguration and continues twisting. Among the policies trained under Stage~1 followed by Stage~2, all five seeds discover contact reconfiguration and continue to accumulate lid rotation over the evaluation horizon.

Figs.~\ref{fig:ab_study}(g) and (h) show representative contact-reconfiguration behaviors observed in simulation. The twist-and-regrasp behavior in Fig.~\ref{fig:ab_study}(g) is observed under both training conditions. The asynchronous finger gaiting in Fig.~\ref{fig:ab_study}(h) is observed only after training under Stage~1 followed by Stage~2 among the evaluated seeds.

\section{Conclusion}

Grasp2Twist combines a continuous enclosure measure, a binary enclosure indicator and a three-stage curriculum to learn a unified policy for bimanual dexterous jar opening. The policy transfers zero-shot to hardware, exhibits asynchronous finger gaiting, and achieves 88\% success across six household containers. Ablations support the roles of enclosure measure and indicator in grasp formation and the curriculum in contact-reconfiguration exploration.

Limitations include less reliable grasp acquisition on smaller containers, potentially affected by state estimation errors; simplified thread mechanics and lid disengagement in simulation; and the current measure and indicator being formulated for three-fingered hand. Future work will address these limitations and extend the approach to other hand morphologies, object geometries, and manipulation tasks.

\section*{Acknowledgment}
ChatGPT was used to convert a real-world robot image into the stylized robot illustration shown in Fig.~\ref{fig:geom_representation}(a).

\addtolength{\textheight}{0 cm}   







\bibliographystyle{IEEEtran}
\bibliography{ref}

\end{document}